\documentclass[10pt,twocolumn,letterpaper]{article}

\usepackage{iccv}              
\usepackage{multirow} 
\usepackage{multicol}  
\usepackage{lipsum}    

\definecolor{iccvblue}{rgb}{0.21,0.49,0.74}
\usepackage[pagebackref,breaklinks,colorlinks,allcolors=iccvblue]{hyperref}

\def\paperID{14431} 
\def\confName{ICCV}
\def\confYear{2025}

\title{Harnessing Text-to-Image Diffusion Models for Point Cloud \\ Self-Supervised Learning}

\author{
Yiyang Chen\textsuperscript{1}\quad 
Shanshan Zhao\textsuperscript{2}\footnotemark[2]\quad
Lunhao Duan\textsuperscript{2}\quad
Changxing Ding\textsuperscript{1,3}\footnotemark[2]\quad
Dacheng Tao\textsuperscript{4}\\
\textsuperscript{1}South China University of Technology \\
\textsuperscript{2}Alibaba International Digital Commerce Group\quad 
\textsuperscript{3}Pazhou Lab\quad 
\textsuperscript{4}Nanyang Technological University\\
    {\tt\small \{eeyiyangchen, sshan.zhao00, dacheng.tao\}@gmail.com, lhduan@whu.edu.cn, chxding@scut.edu.cn}
}

\begin{document}
\maketitle
\renewcommand{\thefootnote}{\fnsymbol{footnote}} 
\footnotetext[2]{Correspondence author.}
\begin{abstract}
Diffusion-based models, widely used in text-to-image generation, have proven effective in 2D representation learning. Recently, this framework has been extended to 3D self-supervised learning by constructing a conditional point generator for enhancing 3D representations. 
However, its performance remains constrained by the 3D diffusion model, which is trained on the available 3D datasets with limited size.
We hypothesize that the robust capabilities of text-to-image diffusion models, particularly  
Stable Diffusion (SD), which is trained on large-scale datasets, can help overcome these limitations. To investigate this hypothesis, we propose PointSD, a framework that leverages the SD model for 3D self-supervised learning. By replacing the SD model's text encoder with a 3D encoder, we train a point-to-image diffusion model that allows point clouds to guide the denoising of rendered noisy images.
With the trained point-to-image diffusion model, we use noise-free images as the input and point clouds as the condition to extract SD features. 
Next, we train a 3D backbone by aligning its features with these SD features, thereby facilitating direct semantic learning. Comprehensive experiments on downstream point cloud tasks and ablation studies demonstrate that the SD model can enhance point cloud self-supervised learning. Code is publicly available at
\url{https://github.com/wdttt/PointSD}.

\end{abstract}

\begin{figure}
\centering
\includegraphics[width=0.9\linewidth]{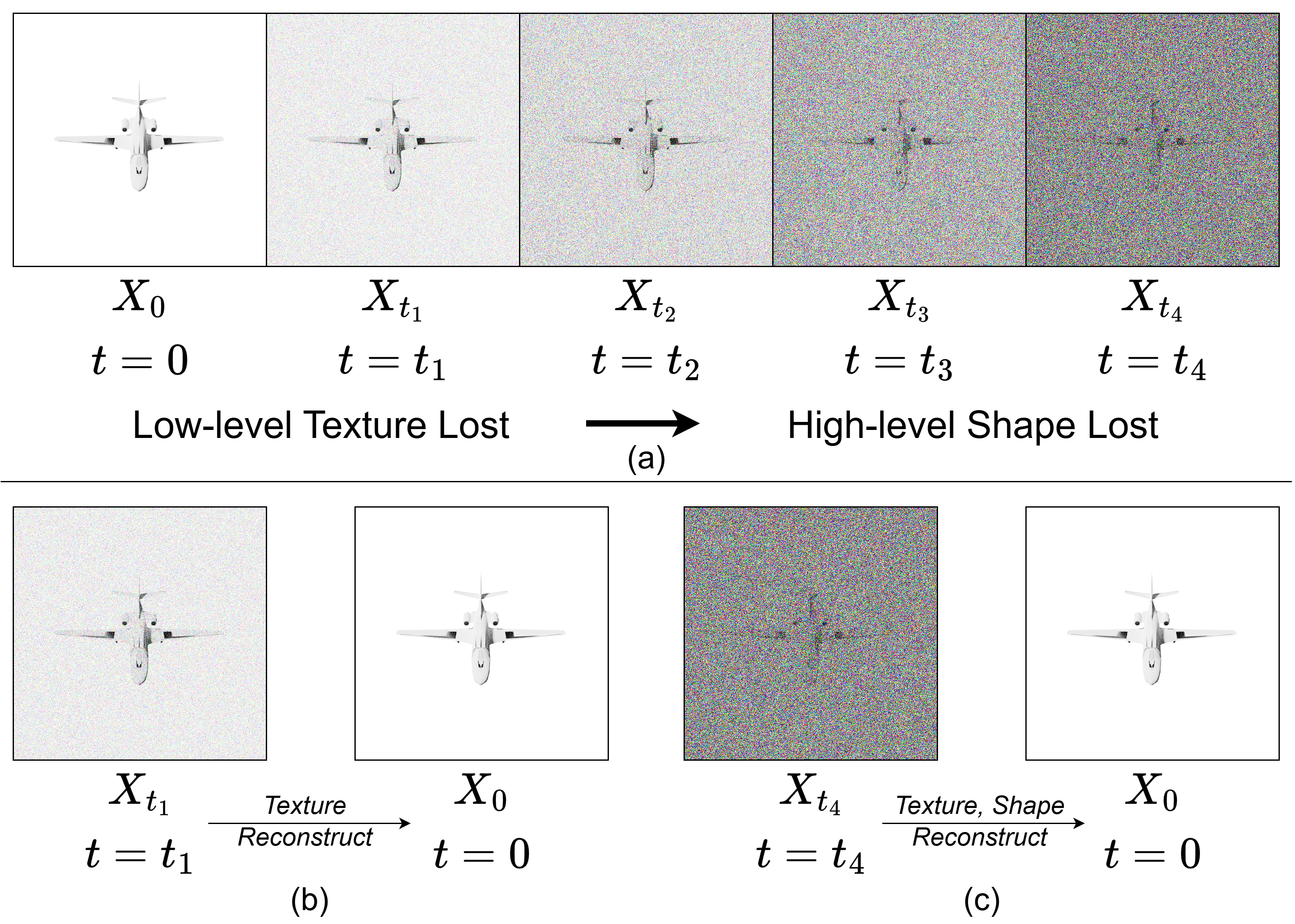}
\vspace{-3mm}
\caption{
Illustration of the forward process and the reverse process of diffusion models. (a) With the increase of noise, the image first loses low-level texture information and then loses high-level shape information. (b) Denoising at low time step $t_1$, recovering texture information. (c) Denoising at high time step $t_4$, recovering texture and shape information. From this figure, we can see that learning via image denoising inevitably introduces low-level information, failing to sufficiently focus on high-level information related to semantics.
}
\label{fig:intro}
\vspace{-6mm}
\end{figure}

\section{Introduction}
\label{sec:intro}

The paradigm of pre-training models through self-supervised learning has been extensively studied in both the vision~\cite{chen2020simple,grill2020bootstrap} and language~\cite{devlin2018bert,brown2020language} domains. Models pre-trained on large amounts of unlabeled data can acquire valuable prior knowledge, resulting in superior performance on downstream tasks compared to training from scratch alone. Numerous effective strategies~\cite{raffel2020exploring,caron2021emerging,he2022masked} have been developed to enhance models' representation capability for pre-training on image or text data.
The success in 2D image and natural language processing has also spurred advancements of self-supervised learning on 3D point cloud data, promoting research in 3D understanding~\cite{qi2017pointnet,qi2017pointnet++,wang2019dynamic,ma2022rethinking,tan2024epmf}.

Current self-supervised learning methods for point clouds can be broadly categorized into two groups: contrastive-based~\cite{afham2022crosspoint,xie2020pointcontrast,zhang2021self} and generative-based~\cite{yu2022point,pang2022masked,wang2023take,Zheng_2024_CVPR}. The core of contrastive methods~\cite{afham2022crosspoint,xie2020pointcontrast,zhang2021self} lies in constructing positive and negative samples, minimizing the distance between positive sample pairs in the embedding space while maximizing the distance between negative pairs. 
Building on the success of generative-based methods such as BERT~\cite{devlin2018bert} for natural language processing and MAE~\cite{he2022masked} for 2D image modeling, Point-BERT~\cite{yu2022point} and Point-MAE~\cite{pang2022masked} integrate these concepts into point cloud self-supervised learning. 
Subsequent studies investigate advanced masking strategies~\cite{Zheng_2024_CVPR,yang2023gd} and enhanced multi-modal interactions~\cite{guo2023joint,chen2023pimae} to further improve performance.

Given the robust capabilities of Diffusion Probabilistic Models (DPMs)~\cite{sohl2015deep,ho2020denoising} in image generation tasks, several works have explored leveraging these models for representation learning in 2D image data~\cite{preechakul2022diffusion,zhang2022unsupervised,yang2023disdiff,yue2024exploring,yang2024diffusion}. While diffusion models have demonstrated significant potential in 2D representation learning, their application to 3D pre-training remains underexplored. To address this, PointDif~\cite{Zheng_2024_CVPR} constructs a diffusion-based conditional point generator and pre-trains a 3D backbone by conditioning on its extracted features, demonstrating the effectiveness of diffusion-based generative models for point cloud pre-training.
However, compared to image and text datasets, the relatively small size of existing 3D datasets limits the learning capacity of the 3D diffusion model, thereby reducing its effectiveness in enabling the 3D backbone to learn robust and generalizable 3D representations.
Text-to-image diffusion models, particularly Stable Diffusion (SD)~\cite{rombach2022high}, have been shown to possess advanced representation capabilities and can be utilized to understand and process downstream 2D understanding tasks~\cite{li2023your, 
zhao2023unleashing, clark2024text, tian2023diffuse}.  
Given its ability to capture rich semantic information in vision understanding tasks, we hypothesize that SD's learned representations can be effectively leveraged to enhance 3D self-supervised learning. By associating point clouds with their corresponding rendered images, we aim to transfer the high-level semantics encoded in SD to the 3D domain.

Building upon this insight, we introduce PointSD, a novel framework that harnesses the frozen SD model to facilitate self-supervised learning for point clouds. 
To leverage the SD model for learning 3D representations, we replace the text encoder with a 3D encoder that takes point clouds as input and train a point-to-image diffusion model based on the SD model.
Specifically, 3D features interact with the intermediate representation of the UNet within the SD model through cross-attention layers, guiding the denoising of the noisy images rendered from the point clouds. 
Although image denoising can assist in learning 3D representations, it inevitably introduces low-level textures and details, 
which may potentially lead the 3D backbone to focus on features irrelevant to semantics, thereby hindering the learning of high-level semantic information. We provide a brief illustration in Fig.~\ref{fig:intro}.

To mitigate this issue and better extract semantic features from the SD model, we draw inspiration from VPD~\cite{zhao2023unleashing}, which effectively captures multi-level semantic representations within SD. VPD achieves this by using noise-free images as input and class-based text prompts as conditions for feature extraction via a denoising step. 
In contrast to VPD, which conditions the SD model on class-based text prompts, we condition our trained point-to-image diffusion model on point clouds to obtain meaningful SD features.
We then train a 3D backbone by aligning the 3D features of an object with its corresponding image features from the SD model.
This alignment allows us to leverage the rich semantics encapsulated in the SD model, enhancing the 3D backbone's ability to understand the object shapes and structures. 

Our main contributions can be summarized as:
(1) Motivated by previous works, this paper studies how to harness the SD model to enhance point cloud self-supervised learning.
(2) We develop PointSD, a framework that leverages existing SD models to assist point cloud pre-training in two stages, where a point-to-image framework is built to extract features from SD, and then feature alignment is performed to encourage the 3D backbone to learn robust representations. 
(3) We conduct experiments to demonstrate that our pre-trained model achieves competitive performance across various downstream tasks.
\section{Related Works}

\subsection{Pre-training for 3D point cloud}
The current point cloud pre-training methods mainly fall into two categories: contrastive-based and generative-based. 
Contrastive-based methods~\cite{afham2022crosspoint,xie2020pointcontrast,zhang2021self} pre-train the backbone by constructing positive and negative sample pairs. For a given sample, the negative sample is usually selected from the remaining samples in the current batch while the positive sample is obtained through data augmentation~\cite{xie2020pointcontrast,zhang2021self} or multi-modal data~\cite{afham2022crosspoint}.
In contrast, 
recovering masked information is the core of generative-based pre-training methods~\cite{yu2022point,pang2022masked,dong2023act,qi2023recon,zhang2023learning,zha2023towards,Zheng_2024_CVPR}.
BERT~\cite{devlin2018bert} and MAE~\cite{he2022masked} methods have achieved success in language and image pre-training tasks, respectively.  
Further, based on the structure of the existing conditional diffusion model, 
PointDif~\cite{Zheng_2024_CVPR} constructs a conditional point-to-point generator and pre-trains the 3D backbone by setting the extracted features as the condition.

Additionally, pre-trained models from other modalities, such as CLIP, have also been explored to address the insufficiency of 3D data.
ULIP~\cite{xue2023ulip} and ULIP-2~\cite{xue2024ulip} align point cloud, image, and text embeddings via the CLIP image model and text model.
ACT~\cite{dong2023act} proposes a teacher-student framework where a CLIP-based cross-modal teacher guides the optimization of 3D student models. 
Meanwhile, ReCon~\cite{qi2023recon} investigates the integration of CLIP's distilled knowledge with self-supervised learning, while I2P-MAE~\cite{zhang2023learning} advances this direction by employing CLIP to inform masking strategies in masked autoencoder training. 
In this paper, we also explore the potential of pre-trained models from other modalities for point cloud pre-training. However, differing from those CLIP-based works~\cite{dong2023act,qi2023recon,zhang2023learning,xue2023ulip,xue2024ulip}, our motivation is to leverage the SD model, which is already pre-trained on large-scale datasets, to enhance 3D representation learning. 
In addition, another distinction of our approach from ULIP, ULIP-2, and ReCon lies in the absence of category labels during training, which is achieved by modifying a pre-trained text-to-image diffusion model to generate images from 3D point clouds.

\subsection{Diffusion Probabilistic Models}
DPMs have made significant progress in the field of image generation~\cite{sohl2015deep,ho2020denoising}.
Currently, many text-to-image diffusion models, such as the SD model~\cite{rombach2022high}, have achieved controllable high-quality image generation. 
The advanced generative capability of the SD model has enabled its incorporation into various generation tasks, such as video generation~\cite{chen2023videocrafter1,blattmann2023stable,guo2023animatediff} and 3D object generation~\cite{liu2023zero,shi2023mvdream,long2023wonder3d}. 

Moreover, the semantic information captured by the SD model has been effectively applied in various downstream tasks, including classification~\cite{clark2024text, li2023your}, semantic segmentation~\cite{tian2023diffuse}, and depth estimation~\cite{zhao2023unleashing}. \cite{li2023your} and \cite{clark2024text} propose that SD can function as a zero-shot classifier, determining the image's category by comparing the difference between the model's predicted noise and the true value under different class-based text prompts. DiffSeg~\cite{tian2023diffuse} performs image segmentation by iteratively combining self-attention maps generated by the SD model across multiple iterations. VPD~\cite{zhao2023unleashing} fuses cross-attention maps with feature maps and forwards them to the decoder for downstream 2D perception tasks.
Our paper also aims to leverage the semantics encapsulated in SD features. While previous works have applied SD to 2D downstream tasks, we extend its application to assist point cloud self-supervised learning.

\begin{figure*}[htbp]
  \centering
  \includegraphics[width=0.9\linewidth]{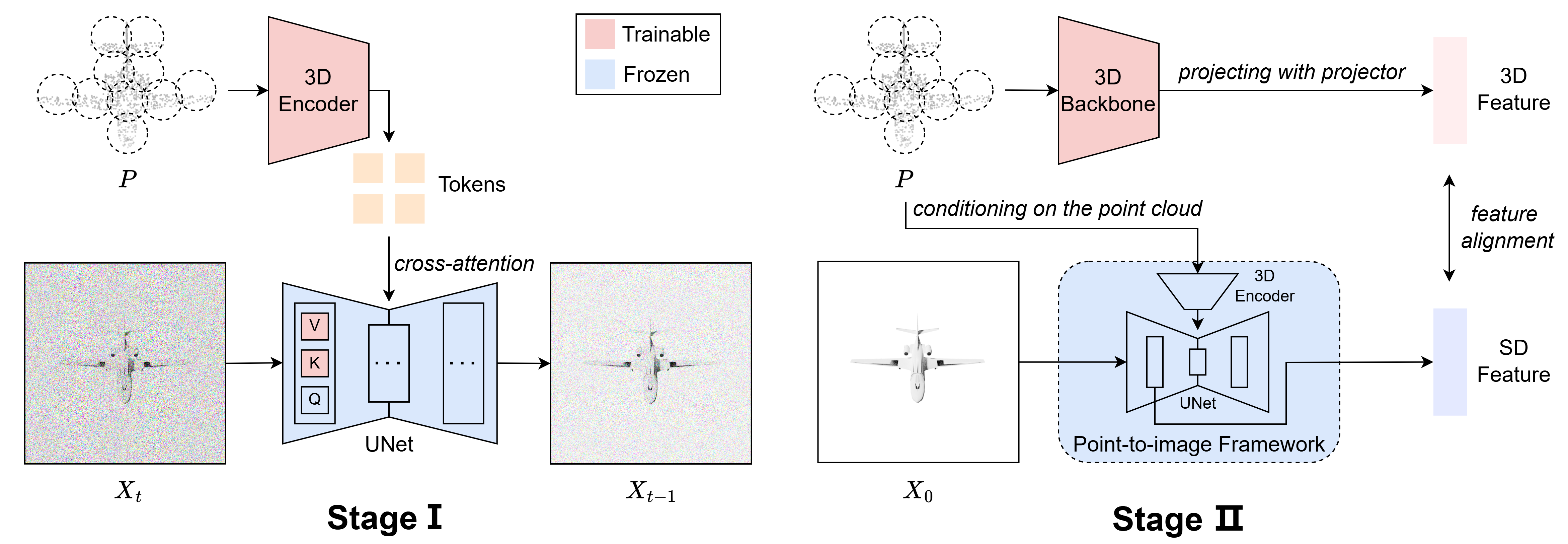} 
      \caption{Overall framework of our PointSD. In the first stage, the 3D encoder extracts features to serve as the condition for guiding the denoising process, forming a point-to-image diffusion framework. 
      In the second stage, feature alignment is performed between 3D features and SD features to boost 3D representation learning.
      }
      \label{fig:framework}
\vspace{-5mm}
\end{figure*}

\section{Methods}
Our framework, PointSD, is shown in Fig.~\ref{fig:framework}. In the
following, we first review some preliminaries about diffusion models and 3D self-supervised learning in Sec.~\ref{sec:preliminary}, and then detail how PointSD leverages the SD model to help the 3D backbone learn strong representations in Sec.~\ref{sec:PointSD}.
\subsection{Preliminary}
\label{sec:preliminary}
\noindent{\bf Diffusion Model.} 
Diffusion models~\cite{ho2020denoising} are probabilistic generative models that define two processes: the forward process and the reverse process.
The forward process is set to a Markov chain of $T$ steps, gradually corrupting the data via Gaussian noise with a variance schedule, while the reverse process involves leveraging a learnable model to generate samples from Gaussian noise. LDM~\cite{rombach2022high} introduces a conditioning mechanism to control the generation of the reverse process. A typical training objective of the diffusion model can be written as:
\begin{equation}
    \begin{aligned}
         L_{denoising} = \mathbb{E}_{X_0,\epsilon\sim\mathcal{N}(0,1),t,c}
    \Vert \epsilon - \epsilon_\theta(X_t,t,c) \Vert_2^2,
    \end{aligned}
    \label{eq:diff}
\end{equation}
where $X_0$ represents the original sample, $t\in[0,T]$ denotes the time step of the forward process, and $X_t$ is the noisy data at $t$ time step. $\epsilon$ is the Gaussian noise, while $\epsilon_\theta$ is the learnable model. $c$ represents an additional condition, such as text or image. 

\noindent{\bf 3D Self-supervised Learning.} 
The objective of 3D self-supervised learning is to learn strong representations that generalize well to downstream tasks. Given a 3D backbone $f$ and a point cloud $P$, 
the representation $R_P$ extracted by the 3D backbone $f$ can be expressed as:
\begin{equation}
    \begin{aligned}
        R_P = f(P).
    \end{aligned}
\end{equation}
For contrastive-based methods~\cite{afham2022crosspoint,xie2020pointcontrast,zhang2021self}, $R_P$ is optimized by comparing it with representations of positive and negative samples.
In comparison, 
generative-based methods~\cite{pang2022masked,zhang2022point} optimize $R_P$ by sending it to a decoder to reconstruct the original sample using a reconstruction loss.

\subsection{PointSD}
\label{sec:PointSD}
PointSD adopts a two-stage training strategy to leverage the semantics encapsulated in the SD model to assist in point cloud pre-training. First, PointSD feeds noisy images rendered from point clouds into a frozen SD model and replaces the text encoder with a 3D encoder to obtain the 3D condition for denoising, thus establishing a point-to-image framework for subsequent feature alignment. Next, PointSD uses point clouds as the condition to extract features from the SD model and then aligns the output of the 3D backbone with them for pre-training.

In the following sections, we separately introduce the construction of the point-to-image framework and the feature alignment operation.
For clarity, we specify that the 3D encoder $g$ is trained in the first stage to construct the point-to-image framework, while the 3D backbone 
$f$ is trained in the second stage to learn strong representations.

\noindent{\bf Stage I: Text-to-image $\Rightarrow$ Point-to-image.}
PointDif~\cite{Zheng_2024_CVPR} introduces an effective approach for utilizing diffusion models in 3D self-supervised learning by conditioning the denoising process on 3D features. However, the relatively small size of existing 3D datasets hinders the training of 3D diffusion models, which in turn limits the ability to learn robust 3D representations.
SD~\cite{rombach2022high}, a text-to-image diffusion model, exhibits advanced representation capabilities and has been successfully applied to a range of downstream 2D understanding tasks~\cite{li2023your, zhao2023unleashing, clark2024text, tian2023diffuse}. Owing to its proficiency in capturing rich semantic information in understanding tasks, we attempt to address data insufficiency using the SD model. 

In our approach, we train a 3D encoder 
$g$ to extract point cloud features and project them into a series of tokens $H$:
\begin{equation}
    \begin{aligned}
        H = g(P),
    \end{aligned}
\end{equation}
where $H$ serves as the conditioning input for the SD model, 
replacing the text features typically provided by the CLIP text encoder to control image generation via the cross-attention mechanism.

The cross-attention mechanism has been widely used in text-to-image models to integrate conditional information into the generation process.
For example, in the cross-attention layers of the SD model, the text features from the CLIP text encoder interact with the intermediate representation of the UNet. The cross-attention mechanism can be formulated as:
\begin{equation}
    \begin{aligned}
        &Attention(Q,K,V) = softmax(\frac{QK^{T}}{\sqrt{d}})V, \\
        &Q = zW_{q}, K = cW_{k}, V = cW_{v},
    \end{aligned}
\end{equation}
where $Q$, $K$, and $V$ are query, key, and value matrices, respectively. $W_q$, $W_k$, and $W_v$ are the corresponding learnable projection matrices. $c$ means the condition, while $z$ denotes the intermediate representation of the UNet. $d$ is the scaling factor.
In our work, we set $H$ as the condition to get new key matrix $K = HW_{k}$ and value matrix $V = HW_{v}$. During training, we only optimize the learnable projection matrices $W_k$ and $W_v$, while keeping $W_q$ fixed.

While this stage helps learn 3D representations, image denoising inevitably introduces low-level textures and details. These elements could distract the 3D encoder, causing it to focus on non-semantic features and hindering the learning of more robust 3D representations. 
To address this limitation, we propose a refinement that better leverages the semantic information encapsulated in the SD model, as detailed in the next section.

\noindent{\bf Stage II: Point Cloud Learning via Feature Alignment.}
Previous work, VPD~\cite{zhao2023unleashing}, presents an effective way to extract semantics from the SD model. 
Unlike the typical diffusion process, VPD uses class-based text prompts as the condition and inputs natural images into the SD model without adding noise, extracting features from the UNet's up-sampling layers for downstream fine-tuning.

Instead of using class-based text prompts as in VPD, we condition our trained point-to-image diffusion model on point clouds, allowing us to extract meaningful SD features for 3D representation learning. While VPD selects features from the up-sampling layers, our experiments show that features from the down-sampling layers lead to better performance in 3D self-supervised learning. 
To leverage these features, we first project the point cloud representation $R_P$ from the 3D backbone $f$ to the same feature space as the SD features using a projector $s$. Then, we perform feature alignment via L2 loss:
\begin{equation}
    \begin{aligned}
       L_{alignment} = \| s(R_P) - R_X \|_2^2,
    \end{aligned}
    \label{eq:align}
\end{equation}
where $R_X$ denotes the 
representation from the last down-sampling layer of the SD model, with the noise-free rendered image $X$ as input and tokens $H$ extracted from corresponding point cloud  $P$ using encoder $g$ as the condition.
Such an alignment encourages the 3D backbone to learn more robust and meaningful features.

\noindent{\bf Training Objective.}
During pre-training, we introduce an augmentation strategy to enhance 3D self-supervised learning.
Recently, mixing-based augmentation strategies~\cite{chen2020pointmixup,lee2021regularization,wang2024pointpatchmix} have been shown to effectively enhance model performance in point cloud tasks by increasing the diversity of training samples. Typically, for two point cloud samples $P$ and $P^{'}$, the mixed sample is obtained by:
\begin{equation}
    \begin{aligned}
       \tilde{P} = M \odot P + (1-M) \odot P^{'},
    \end{aligned}
\end{equation}
where $M$ denotes a binary mask indicating which point needs to be retained to construct the mixed point cloud,
and $\odot$ represents element-wise multiplication. In the implementation, we first divide point clouds into patches and then mix them at the patch level.
Further, we build the augmented image sample to pair with $\tilde{P}$ as follows:
\begin{equation}
    \begin{aligned}
       \tilde{X} = X \cup X^{'},
    \end{aligned}
\end{equation}
where $X$ and $X^{'}$ are both image samples, and $\cup$ represents stitching two images along the width. 
With augmented input-condition sample pairs, the 3D backbone will be forced to learn complex representations from $\tilde{P}$. We apply this augmentation strategy in both training stages. 

For the first stage, only 3D encoder $g$ and projection matrices $W_k$ and $W_v$ are trainable, while the remaining parameters of the UNet in the SD model are frozen.
We set projected tokens $H$ from the point cloud as the condition, and the Eq.~\ref{eq:diff} can be rewritten as:

\begin{equation}
\resizebox{\linewidth}{!}{
    $L_{denoising} = \left \{ \begin{aligned}
    &\mathbb{E}_{\tilde{X}_0,\epsilon\sim\mathcal{N}(0,1),t,\tilde{H}} 
    \Vert \epsilon - \epsilon_\theta(\tilde{X}_t,t,\tilde{H}) \Vert_2^2, \quad
    &\alpha>0.5,\\
    &\mathbb{E}_{X_0,\epsilon\sim\mathcal{N}(0,1),t,H} 
    \Vert \epsilon - \epsilon_\theta(X_t,t,H) \Vert_2^2, \quad
    &\alpha\leq0.5,
    \end{aligned} \right.$
    }
\end{equation}
where $\alpha$ is sampled from the uniform distribution $\mathcal{U}(0,1)$. 
$X_0$ and $\tilde{X}_0$ are the original rendered image and augmented image (two rendered images are stitched), while $X_t$ and $\tilde{X}_t$ represent adding noise on both images at $t$ time step.
$\tilde{H}$ denotes the projected tokens from augmented point cloud $\tilde{P}$. We use augmented samples in training with a 50\% probability to prevent over-augmentation, which would otherwise be detrimental to performance.

For the second stage, we only train the 3D backbone $f$ and the projector $s$, while the parameters of the SD model and the 3D encoder $g$ are frozen.
We perform feature alignment, and with the augmentation strategy, Eq.~\ref{eq:align} can be rewritten as:

\begin{equation}
   L_{alignment} = \left\{
   \begin{aligned}
   &\Vert s(R_{\tilde{P}}) - R_{\tilde{X}} \Vert_2^2, \quad 
   & \alpha > 0.5, \\
   &\Vert s(R_P) - R_X \Vert_2^2, \quad 
   & \alpha \leq 0.5,
   \end{aligned}
   \right.
\end{equation}
where $R_{\tilde{P}}$ denotes the 3D representation of augmented point cloud $\tilde{P}$ and $R_{\tilde{X}}$ means the 2D representation of augmented rendered image $\tilde{X}$ from SD model.

By minimizing the above losses, the 3D backbone will be encouraged to capture the differences between point cloud samples and learn discriminative representations.

\section{Experiments}
In this section, we first introduce our pre-training experiments in Sec.~\ref{sec:pre-train}. Next, in Sec.~\ref{sec:downstream} we evaluate our pre-trained model on downstream tasks, including object classification, few-shot learning, part segmentation, and object detection. Moreover, we provide comprehensive ablation studies in Sec.~\ref{sec:ablation} to further analyze our PointSD.

\subsection{Self-supervised Pre-training}
\label{sec:pre-train}
\noindent{\bf Setting.}
We perform self-supervised pre-training on the ShapeNet~\cite{chang2015shapenet} dataset, which contains approximately 52,470 samples across 55 categories. 
For each 3D object, we follow prior work~\cite{yu2022point,pang2022masked} to sample 1,024 points and divide them into 64 groups. Each group is obtained by the KNN algorithm, with each group containing 32 points. Additionally, to ensure a fair comparison with mainstream self-supervised learning works~\cite{yu2022point,pang2022masked}, we adopt the typical transformer architecture as the 3D backbone, setting the embedding dimension to 384 and configuring the number of blocks and heads to 12 and 6, respectively. Our 3D encoder adopts the same architecture as the 3D backbone. Our projector also utilizes a transformer architecture, with an embedding dimension set to 768, and the number of blocks and attention heads configured to 3 and 12, respectively. 
For rendered images, we use the rendered data provided in ULIP~\cite{xue2023ulip}, where each 3D object in the ShapeNet dataset is rendered into 30 RGB images with evenly spaced views (one every 12 degrees).
For the pre-trained SD model, we use SD v1.5 for our experiments.

\begin{table}[t]
\centering
\caption{Classification accuracy (\%) on the three subsets of ScanObjectNN dataset and ModelNet40 dataset. \#Params (M) and \#FLOPs (G) represent the number of model parameters and floating point operations, and Time (h) refers to fine-tuning time on the
PB-T50-RS setting using a single NVIDIA RTX 4090 GPU. 
$*$ denotes using class labels during pre-training.}
\vspace{-2mm}
\label{tab:cls}
\resizebox{\linewidth}{!}{
\begin{tabular}{lccccccc}
\hline
\multirow{2}{*}{Method} & \multicolumn{3}{c}{Efficiency} & \multicolumn{3}{c}{ScanObjectNN} & \multirow{2}{*}{ModelNet40} \\ \cline{2-7} 
                        & \#Params   & \#FLOPs   & Time  & OBJ-BG  & OBJ-ONLY  & PB-T50-RS  \\ \hline          
\multicolumn{8}{c}{\textit{Supervised Learning Only}}                                                                                    \\ \hline
PointNet~\cite{qi2017pointnet}             & 3.5        & -  & -              & 73.3    & 79.2      & 68.0       & 89.2                                            \\
PointNet++~\cite{qi2017pointnet++}              & 1.5    & -  & -                    & 82.3    & 84.3      & 77.9       & 90.7                                            \\
Transformer~\cite{yu2022point}             & 22.1       & -  & -                 & 79.9   & 80.6     & 77.2      & 91.4                                            \\
DGCNN~\cite{wang2019dynamic}                   & 1.8   & -  & -                      & 82.8    & 86.2      & 78.1       & 92.9                                            \\
PointCNN~\cite{NEURIPS2018_f5f8590c}                & 0.6   & -  & -  
& 86.1    & 85.5      & 78.5      & 92.2                                            \\
SimpleView~\cite{pmlr-v139-goyal21a}              & -  & -  & -                        & -       & -         & 80.5       & 93.9                                            \\
MVTN~\cite{Hamdi_2021_ICCV}                    & 11.2    & -  & -                    & -       & -         & 82.8       & 93.8                                            \\
PointMLP~\cite{ma2022rethinking}                & 12.6    & -  & -                    & -       & -         & 85.4 $\pm$ 0.3   & \textbf{94.1}                                            \\
PointNeXt~\cite{NEURIPS2022_9318763d}               & 1.4   & -  & -                      & -       & -         & 87.7 $\pm$ 0.4   & 93.7                                            \\
P2P-HorNet~\cite{NEURIPS2022_5cd6dc94}              & 195.8  & -  & -                     & -       & -         & 89.3       & 94.0                                            \\ \hline
\multicolumn{8}{c}{\textit{Single-Modal Self-Supervised Learning}}                                                                       \\ \hline
Point-BERT~\cite{yu2022point}              & 22.1   & 4.8 & 2.0                   & 87.43   & 88.12     & 83.07      & 92.7                                            \\
MaskPoint~\cite{liu2022masked}               & 22.1   & 4.8 & 2.0                   & 89.30   & 88.10     & 84.30      & -                                               \\
Point-MAE~\cite{pang2022masked}                & 22.1   & 4.8  &  2.0                  & 90.02   & 88.29     & 85.18      & 93.2                                            \\
Point-M2AE~\cite{zhang2022point}               & 15.3   & 3.6 & 5.0                   & 91.22   & 88.81     & 86.43      & 93.4                                            \\
PointDif~\cite{Zheng_2024_CVPR}                & 22.1   & 4.8 & 2.0                      & 93.29   & 91.91     & 87.61      & -                                               \\
Point-FEMAE~\cite{zha2023towards}              & 27.4  & 14.2 & 3.1                    & \textbf{95.18}   & 93.29     & 90.22      & 94.0                                            \\ 
\hline
\multicolumn{8}{c}{\textit{Cross-Modal Self-Supervised Learning}}                                                                        \\ \hline
ACT~\cite{dong2023act}                     & 22.1   & 4.8 & 2.0                   & 93.29   & 91.91     & 88.21      & 93.2                                            \\
Joint-MAE~\cite{guo2023joint}               & -  & - & -
  & 90.94   & 88.86     & 86.07      & -
                                     \\
I2P-MAE~\cite{zhang2023learning}                 & 15.3  & 3.6 & 5.0                    & 94.15   & 91.57     & 90.11      & 93.7                                            \\
ReCon$^{*}$~\cite{qi2023recon}                   & 43.6   & 5.3 & 7.0                   & \textbf{95.18}   & \textbf{93.63}     & \textbf{90.63}      & \textbf{94.1}                                            \\
TAP~\cite{wang2023take}                     & 22.1    & 4.8 & 2.0                  & 90.36   & 89.50     & 85.67      & -                                               \\
Ours       & 22.1   & 4.8 & 2.0                   & \textbf{95.18}   & \textbf{93.63}     & 90.08      & 93.7                                            \\
\hline
\end{tabular}
}
\vspace{-5mm}
\end{table}

\noindent{\bf Training Scheme.} For both training stages, we use the AdamW~\cite{loshchilov2017decoupled} optimizer and adjust the learning rate with a cosine scheduler~\cite{loshchilov2016sgdr}.
The initial learning rate and total batch size are set to 0.000125 and 128, respectively. We train the model on 2 NVIDIA RTX 4090 GPUs (24G memory each), and training takes approximately 18 hours.
The maximum number of training epochs is set to 300, with the first 10 epochs for warm-up. The $t$ for the diffusion process is randomly sampled from $[500,1000]$ in the first stage. Furthermore, the second stage is trained from scratch rather than fine-tuned from the first stage weights.

\subsection{Downstream Tasks}
\label{sec:downstream}

For all downstream tasks, following previous works~\cite{pang2022masked,yu2022point}, we only keep the pre-trained 3D backbone and add a prediction head on it. All parameters of both the pre-trained 3D backbone and the newly added head are updated.

\begin{table}[t]
\centering
\caption{Few-shot learning results on ModelNet40 dataset. The results on the left and right sides of $\pm$ are the average accuracy (\%) and standard deviation obtained by 10 independent experiments. $*$ denotes using class labels during pre-training.}
\vspace{-2mm}
\label{tab:fewshot}
\resizebox{\linewidth}{!}{
\begin{tabular}{lllll}
\hline
\multirow{2}{*}{Method} & \multicolumn{2}{c}{5-way}                                 & \multicolumn{2}{c}{10-way}                                \\ \cline{2-5} 
                        & \multicolumn{1}{c}{10-shot} & \multicolumn{1}{c}{20-shot} & \multicolumn{1}{c}{10-shot} & \multicolumn{1}{c}{20-shot} \\ \hline
\multicolumn{5}{c}{\textit{Supervised Learning Only}}                                                                                           \\ \hline
DGCNN~\cite{wang2019dynamic}                   & 31.6 $\pm$ 2.8                  & 40.8 $\pm$ 4.6                  & 19.9 $\pm$ 2.1                  & 16.9 $\pm$ 1.5                  \\
Transformer~\cite{yu2022point}             & 87.8 $\pm$ 5.2                  & 93.3 $\pm$ 4.3                  & 84.6 $\pm$ 5.5                  & 89.4 $\pm$ 6.3                  \\
\hline
\multicolumn{5}{c}{\textit{Single-Modal Self-Supervised Learning}}                                                                              \\ \hline
DGCNN-OcCo~\cite{wang2019dynamic}
& 90.6 $\pm$ 2.8                  & 92.5 $\pm$ 1.9                  & 82.9 $\pm$ 1.3                  & 86.5 $\pm$ 2.2                  \\
Transformer-OcCo~\cite{yu2022point}        & 94.0 $\pm$ 3.6                  & 95.9 $\pm$ 2.3                  & 89.4 $\pm$ 5.1                  & 92.4 $\pm$ 4.6                  \\
Point-BERT~\cite{yu2022point}              & 94.6 $\pm$ 3.1                  & 96.3 $\pm$ 2.7                  & 91.0 $\pm$ 5.4                  & 92.7 $\pm$ 5.1                  \\
MaskPoint~\cite{liu2022masked}               & 95.0 $\pm$ 3.7                  & 97.2 $\pm$ 1.7                  & 91.4 $\pm$ 4.0                  & 93.4 $\pm$ 3.5                  \\
Point-MAE~\cite{pang2022masked}               & 96.3 $\pm$ 2.5                  & 97.8 $\pm$ 1.8                  & 92.6 $\pm$ 4.1                  & 95.0 $\pm$ 3.0                  \\ 
Point-M2AE~\cite{zhang2022point}              &  96.8 $\pm$ 1.8                              & 98.3 $\pm$ 1.4                            & 92.3 $\pm$ 4.5                            & 95.0 $\pm$ 3.0                            \\
Point-FEMAE~\cite{zha2023towards}             & 97.2 $\pm$ 1.9                               & 98.6 $\pm$ 1.3                            & \textbf{94.0 $\pm$ 3.3}                            & 95.8 $\pm$ 2.8                            \\ 
\hline
\multicolumn{5}{c}{\textit{Cross-Modal Self-Supervised Learning}}                                                                               \\ \hline
ACT~\cite{dong2023act}                     & 96.8 $\pm$ 2.3                               & 98.0 $\pm$ 1.4                            & 93.3 $\pm$ 4.0                            & 95.6 $\pm$ 2.8                            \\
Joint-MAE~\cite{guo2023joint}               &  96.7 $\pm$ 2.2                              & 97.9 $\pm$ 1.8                            & 92.6 $\pm$ 3.7                             & 95.1 $\pm$ 2.6                            \\
I2P-MAE~\cite{zhang2023learning}                 & 97.0 $\pm$ 1.8                              & 98.3 $\pm$ 1.3                            & 92.6 $\pm$ 5.0                            & 95.5 $\pm$ 3.0                            \\
ReCon$^{*}$~\cite{qi2023recon}                   & 97.3 $\pm$ 1.9                                & 98.9 $\pm$ 1.2                            & 93.3 $\pm$ 3.9                            & 95.8 $\pm$ 3.0                            \\
TAP~\cite{wang2023take}                   & 97.3 $\pm$ 1.8                               & 97.8 $\pm$ 1.7                            & 93.1 $\pm$ 2.6                             & 95.8 $\pm$ 1.0                            \\
Ours       & \textbf{97.7 $\pm$ 1.8}                            &  \textbf{99.0 $\pm$ 0.9}                            & 93.8 $\pm$ 3.6                            & \textbf{95.9 $\pm$ 2.6}                            \\
\hline
\end{tabular}
}
\vspace{-5mm}
\end{table}

\begin{table}[t]\small
\centering
\caption{Part segmentation results on ShapeNetPart dataset. mIoU$_{I}$ (\%) and mIoU$_{C}$ (\%)
denote mean IoU over all instances and all classes respectively. $*$ denotes using class labels during pre-training.}
\vspace{-2mm}
\label{tab:partseg}
\begin{tabular}{lcc}
\hline
Method                         & mIoU$_{C}$            & mIoU$_{I}$            \\ \hline
\multicolumn{3}{c}{\textit{Supervised Learning Only}}              \\ \hline
DGCNN~\cite{wang2019dynamic}                          & 82.3            & 85.2            \\
PointMLP~\cite{ma2022rethinking}                       & 84.6            & 86.1            \\
Transformer~\cite{yu2022point}                    & 83.4            & 84.7            \\ \hline
\multicolumn{3}{c}{\textit{Single-Modal Self-Supervised Learning}} \\ \hline
Transformer-OcCo~\cite{yu2022point}               & 83.4            & 85.1            \\
Point-BERT~\cite{yu2022point}                     & 84.1            & 85.6            \\
MaskPoint~\cite{liu2022masked}                      & 84.4            & 86.0            \\
Point-MAE~\cite{pang2022masked}                      & 84.2            & 86.1            \\
Point-M2AE~\cite{zhang2022point}                     & 84.9            & 86.5            \\
Point-FEMAE~\cite{zha2023towards}                    & 84.9            & 86.3            \\ 
\hline
\multicolumn{3}{c}{\textit{Cross-Modal Self-Supervised Learning}}           \\ \hline
ACT~\cite{dong2023act}                            & 84.7            & 86.1            \\
Joint-MAE~\cite{guo2023joint}                      & \textbf{85.4}            & 86.3            \\
I2P-MAE~\cite{zhang2023learning}                        & 85.2            & \textbf{86.8}            \\
ReCon$^{*}$~\cite{qi2023recon}                          & 84.8            & 86.4            \\
Ours             & 84.5            & 86.1            \\
\hline
\end{tabular}
\vspace{-5mm}
\end{table}

\noindent{\bf Object Classification.}
We evaluate the performance of the pre-trained model on classification tasks using the ScanObjectNN~\cite{uy2019revisiting} and ModelNet40~\cite{wu20153d} datasets. 

ScanObjectNN is a real-world dataset containing 15,000 objects from 15 categories. We evaluate the performance of our pre-trained model on three different settings: OBJ-BG, OBJ-ONLY, and PB-T50-RS. Specifically, OBJ-ONLY contains only the object, OBJ-BG adds unsegmented background, and PB-T50-RS further applies 50\% translation, rotation, and scaling to the bounding box.
Previous works~\cite{yu2022point,pang2022masked} fine-tune models on ScanObjectNN dataset with scale and translation augmentations. Recent studies~\cite{qi2023recon, zhang2023learning, zha2023towards} find that applying rotation augmentation during training can effectively improve testing performance.
Therefore, we follow these approaches by using rotation augmentation and sampling 2048 points for each point cloud in our experiments to compare with the state-of-the-art performance. 

As shown in Tab.~\ref{tab:cls}, we categorize existing works into three groups.
To facilitate a comprehensive comparison with self-supervised learning methods, we also report their FLOPs and fine-tuning time.
First, our method surpasses the baseline Transformer~\cite{yu2022point} by 15.28\%, 13.03\%, and 12.88\% in the three ScanObjectNN settings, respectively. Next, among single-modal methods, 
PointDif~\cite{Zheng_2024_CVPR} constructs a conditional point-to-point generator, which trains the 3D backbone via denoising. However, its performance is constrained by the 3D diffusion model, trained on the limited size of available 3D datasets.
In contrast, we leverage the SD model pre-trained on large-scale datasets to enhance 3D self-supervised learning. As shown in Tab.~\ref{tab:cls}, our method consistently outperforms PointDif across all three ScanObjectNN settings.
Furthermore, compared with ReCon~\cite{qi2023recon}, a state-of-the-art cross-modal method that uses class information during pre-training, our approach achieves competitive performance while requiring only about half the number of parameters.

For ModelNet40, following Point-BERT~\cite{yu2022point}, we sample each point cloud to a fixed size of 1,024 points and report accuracy.
Compared to Transformer~\cite{yu2022point}, which is trained from scratch, our method outperforms it by 2.3\%, indicating that our pre-training framework effectively helps the 3D backbone learn meaningful representations. 
Compared to single-modal self-supervised learning methods~\cite{yu2022point, pang2022masked, liu2022masked} that adopt the same backbone as ours, our approach achieves significant performance gains by leveraging the pre-trained SD model. Point-FEMAE~\cite{zha2023towards} introduces a two-branch network to enhance 3D self-supervised learning, achieving higher performance than our method but at the cost of increased model complexity.
Compared to other cross-modal self-supervised learning methods~\cite{dong2023act, zhang2023learning, qi2023recon} based on CLIP, our method still performs competitively. 

\begin{table}[t]\small
\caption{Object detection results on ScanNetV2 dataset. AP$_{0.5}$ represents average precision at an IoU threshold of 0.5 for detection. }
\vspace{-2mm}
\label{tab:detection}
\centering
\begin{tabular}{lcc}
\hline
Method     & Pre-training Dataset & AP$_{0.5}$   \\ \hline
3DETR~\cite{misra2021end}      & -                 & 37.9 \\
Point-BERT~\cite{yu2022point} & ScanNet-Medium    & 38.3 \\
MaskPoint~\cite{liu2022masked}  & ScanNet-Medium    & 40.6 \\
TAP~\cite{wang2023take}        & ShapeNet          & 41.4 \\
Ours       & ShapeNet          & \textbf{42.4} \\ \hline
\end{tabular}
\vspace{-5mm}
\end{table}

\begin{table*}[t]\small
\centering
\begin{minipage}[t!]{0.3\linewidth}
\centering
\captionsetup{width=0.8\linewidth}
\caption{Ablation study on conditions.}
\vspace{-2mm}
\label{tab:ablation_cond}
\begin{tabular}{@{}cc@{}}
\toprule
Conditions & ScanObjectNN \\ \midrule
zero      & 89.31 \\
cls      & 89.42 \\
pc      & 90.08 \\ \bottomrule
\end{tabular}
\end{minipage}
\begin{minipage}[t!]{0.3\linewidth}
\centering
\caption{Ablation study on augmentation strategies.}
\vspace{-2mm}
\label{tab:ablation_aug}
\begin{tabular}{@{}cc@{}}
\toprule
Augmentation & ScanObjectNN \\ \midrule
no aug         & 88.86 \\
mix pc          & 89.07 \\
mix pc \& stitch img       & 90.08 \\
 \bottomrule
\end{tabular}
\end{minipage}
\begin{minipage}[t!]{0.3\linewidth}
\centering
\captionsetup{width=0.8\linewidth}
\caption{Ablation study on SD layers.}
\vspace{-2mm}
\label{tab:ablation_layers}
\begin{tabular}{@{}cc@{}}
\toprule
SD Layers & ScanObjectNN \\ \midrule
down         & 86.78 \\
mid     & 87.12 \\ 
up          & 90.08 \\
\bottomrule
\end{tabular}
\end{minipage}
\vspace{-5mm}
\end{table*}
\noindent{\bf Few-shot Learning.}
In addition to fine-tuning with full data, we also evaluate the performance of our method under a few-shot setting on the ModelNet40 dataset. Following previous works~\cite{sharma2020self}, during training, we randomly select $n\in\{5,10\}$ classes from the dataset and $k\in\{10,20\}$ samples for each selected class (i.e., n-way k-shot). In $n$ selected classes, 20 unseen samples will be sampled from each class for testing.
As indicated in Tab.~\ref{tab:fewshot}, our approach demonstrates clear advantages in the 5-way 10-shot, 5-way 20-shot, and 10-way 20-shot settings.

\begin{figure}[t]
  \centering
  \includegraphics[width=0.75\linewidth]{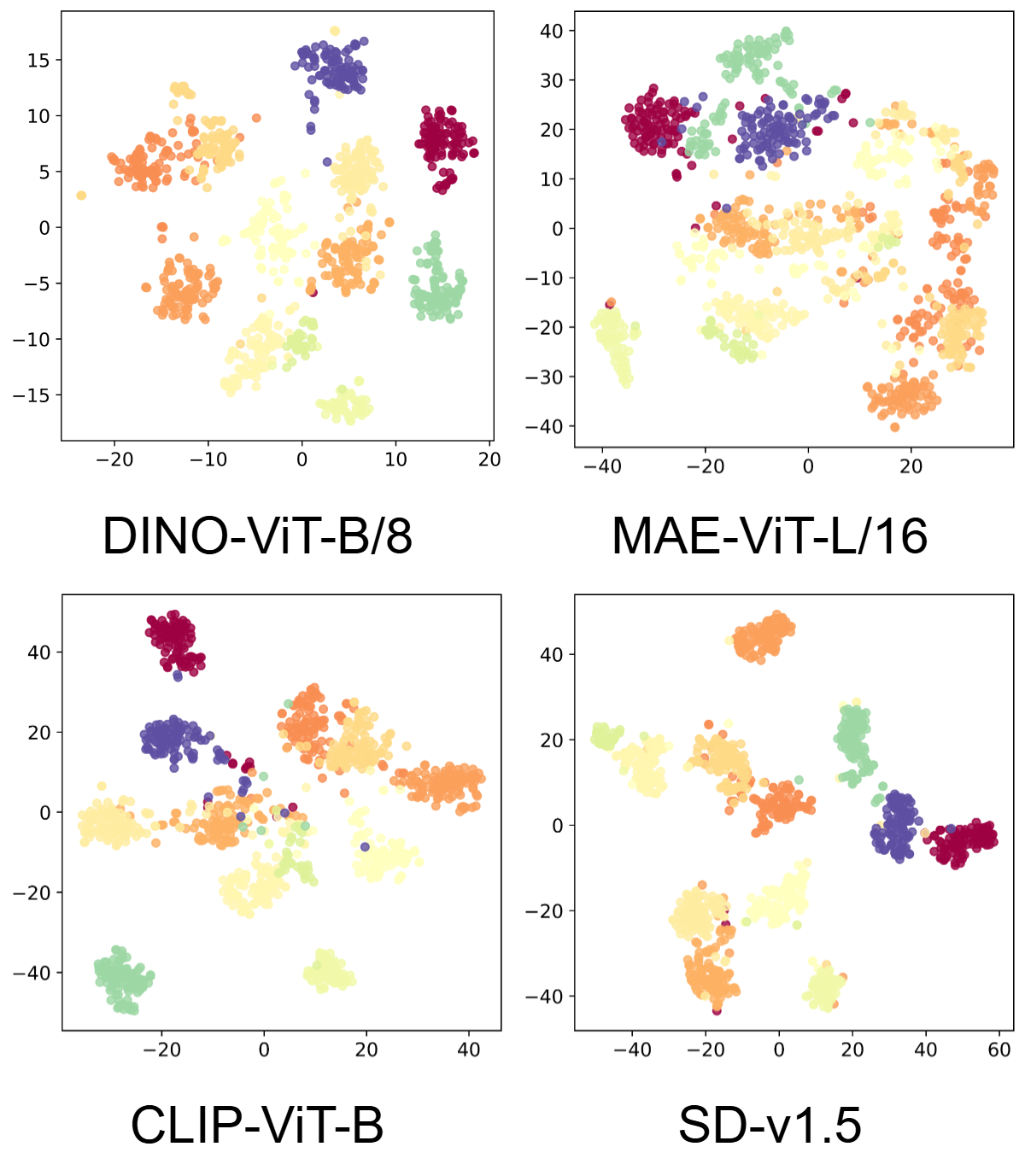} 
      \caption{t-SNE visualization on a subset of the categories from the ShapeNet dataset.}
      \label{fig:tsne2}
      \vspace{-13pt}
\end{figure}

\noindent{\bf Part Segmentation.}
We evaluate the performance of our pre-trained model on the part segmentation task with the ShapeNetPart~\cite{yi2016scalable} dataset, which consists of 16,881 3D samples from 16 categories, further subdivided into 50 parts. We follow PointNet~\cite{qi2017pointnet} to split data and sample 2,048 points for each 3D object. 
Here, we report both mean Intersection-over-Union (mIoU) over all instances and classes for better comparison. 

As shown in Tab.~\ref{tab:partseg}, our pre-training approach improves 1.1\% on mIoU$_{C}$ and 1.4\% on mIoU$_{I}$ compared to baseline Transformer~\cite{yu2022point}. Compared to recent methods, our performance on part segmentation lags behind to some extent. This may be because they train the 3D backbone using patch-level reconstruction loss, whereas we perform alignment with an object-level loss. We guess that, since part segmentation requires point-wise predictions, dense supervision at the patch level contributes to better performance.

\noindent{\bf Object Detection.}
To better evaluate the effectiveness of our method on scene-level prediction tasks, we follow MaskPoint~\cite{liu2022masked} to choose ScanNetV2~\cite{dai2017scannet} with the 3DETR~\cite{misra2021end} backbone to perform experiments. ScanNetV2 is a large-scale 3D dataset containing over 1,500 indoor scenes, annotated with object-level and semantic segmentation labels. 3DETR is a common end-to-end 3D object detection pipeline based on transformer structure. We pre-train 3DETR backbone on the ShapeNet dataset and then fine-tune it on ScanNetV2.

We report the results of different methods based on the 3DETR backbone in Tab.~\ref{tab:detection}. Point-BERT and MaskPoint use a subset of ScanNetV2 as the pre-training dataset, while TAP and our method use ShapeNet for pre-training, which is more challenging due to the domain gap between the two datasets. From Tab.~\ref{tab:detection}, our method brings a 4.5\% improvement compared to the baseline 3DETR and outperforms other pre-training methods.

\subsection{Ablation Studies}
\label{sec:ablation}

We conduct ablation studies on the conditions of the SD model, augmentation strategies, SD layers for feature extraction, pre-trained models for 3D self-supervised learning, and the effect of different training stages.
We perform experiments based on SD v1.5 and the PB-T50-RS setting of the ScanObjectNN dataset and report performance with accuracy (\%).

\begin{table}[t]\small
\caption{Ablation study on pre-trained models.}
\vspace{-2mm}
\label{tab:models}
\centering
\begin{tabular}{lc}
\hline
Pre-trained Models & ScanObjectNN \\ \hline
DINO-ViT-B/8~\cite{caron2021emerging}       & 89.38             \\ 
MAE-ViT-L/16~\cite{he2022masked}       & 88.72             \\ 
CLIP-ViT-B~\cite{radford2021learning}         & 89.35             \\ 
SD-v1.5~\cite{rombach2022high}            & 90.08             \\
\hline
\end{tabular}
\vspace{-5mm}
\end{table}

\noindent{\bf Conditions.}
Considering that our second training stage can also work without the point cloud condition, we conduct experiments to verify the necessity of the first training stage. We perform the second training stage under three conditions: `zero' refers to not using any condition, `cls' refers to using class-based text as the condition, and `pc' refers to using point clouds as the condition. As shown in Tab.~\ref{tab:ablation_cond}, `pc' outperforms the others in terms of performance. Despite the additional label information, `cls' may still suffer from misalignment between text and images, resulting in weaker performance compared to the aligned `pc' condition. This shows that our first training stage effectively enhances feature alignment in the second training stage.

\noindent{\bf SD Layers.}
As shown in Tab.~\ref{tab:ablation_layers}, we investigate the impact of features from different layers of the SD model on performance. The terms `down', `mid', and `up' refer to the down-sampling layers, middle layer, and up-sampling layers, respectively. The best performance is achieved by the features from down-sampling layers, while the worst performance is observed from the up-sampling layers. This can be attributed to the fact that during up-sampling, low-level image information is progressively introduced. In contrast, the down-sampling layers preserve higher-level information more effectively. High-level features are more closely related to semantic information, while low-level features are associated with underlying details. Therefore, the features from down-sampling layers are more beneficial for point cloud self-supervised learning.

\noindent{\bf Pre-trained Models.}
We also replace the SD model with other pre-trained models to evaluate the performance of our method, as shown in Tab.~\ref{tab:models}. 
For a fair comparison, we only use the image model of CLIP in the experiments. The results demonstrate that, compared to other pre-trained models, the SD model contains richer semantic information, which better supports 3D self-supervised learning. 

Additionally, we provide t-SNE~\cite{van2008visualizing} visualizations of features from the pre-trained models listed in Tab.~\ref{tab:models}, using the ShapeNet~\cite{chang2015shapenet} dataset.
Notably, we use point clouds as the condition to extract features from SD-v1.5. To ensure clearer visualizations, we select a subset of categories from the ShapeNet dataset.

As shown in Fig.~\ref{fig:tsne2}, SD-v1.5 demonstrates a superior ability to distinguish features from different classes, outperforming the other pre-trained models. This indicates the potential of enhancing 3D self-supervision by leveraging the semantics encoded in the SD model.

\noindent{\bf Training Stages.}
Here we conduct experiments on the effect of our different training stages. From Tab.~\ref{tab:stages}, `low time step', `normal time step', and `high time step'  refer to selecting a diffusion step $t$ from the intervals [0, 500], [0, 1000], and [500, 1000], respectively, during the first training stage. `w/ alignment' and `w/o alignment' indicate whether the second training stage is applied.

According to the experimental results, when the second stage is not conducted, denoising from high diffusion steps yields better performance, suggesting that the low-level information learned at low diffusion steps may hinder the learning of better 3D representations. However, denoising from high diffusion steps still inevitably introduces certain low-level details, indicating the need for further improvement. Subsequent results show that feature alignment leads to additional performance gains, indicating that aligning 3D features with SD intermediate features helps to better leverage the semantics embedded in the SD model. 
Furthermore, higher performance in the first training stage leads to better feature alignment in the second training stage.

\noindent{\bf Augmentation Strategies.}
We compare different augmentation strategies in Tab.~\ref{tab:ablation_aug}. First, our strategy (`mix pc \& stitch img') builds mixed point clouds and stitched images to introduce augmented training samples, outperforming the baseline (`no aug') by 1.22\%. Further, `mix pc' represents only building mixed point clouds, which is a common augmentation strategy in the 3D domain. However, it yields only a marginal improvement over the baseline. Since the mixed point cloud is composed of two samples, pairing it with the original image may lead the 3D backbone to learn incorrect representations. 

\begin{table}[t]\small
\caption{Ablation study on training stages.}
\vspace{-2mm}
\label{tab:stages}
\centering
\begin{tabular}{lc}
\hline
Method                         & ScanObjectNN \\ \hline
low time step w/o alignment    & 87.30        \\
normal time step w/o alignment & 88.06        \\
high time step w/o alignment   & 88.45        \\
low time step w/ alignment   & 88.58        \\
normal time step w/ alignment   & 89.31        \\
high time step w/ alignment    & 90.08        \\ \hline
\end{tabular}
\vspace{-3.5mm}
\end{table}

\section{Conclusion and Limitations}
This paper aims to harness the SD model for point cloud self-supervised learning.
To achieve this, we propose PointSD, a point cloud pre-training framework that exploits the SD model's ability to enable a 3D backbone to learn representations with strong capacity. 
Specifically, we introduce a two-stage training strategy: in the first stage, a point-to-image framework based on the SD model is established to extract semantically-rich SD features, and in the second stage, feature alignment is performed.
Experimental results on downstream tasks demonstrate that within our framework the SD model effectively helps the 3D backbone learn meaningful representations.

Witnessing the success of generative-based 3D pre-training methods such as Point-BERT and Point-MAE, subsequent works have explored improvements in model design, masking strategies, and multi-modal interactions. Our motivation is to investigate whether and how the SD model can benefit 3D pre-training. Experimental results demonstrate that our method outperforms Point-BERT and Point-MAE but remains only comparable to recent state-of-the-art approaches. 
Hence, incorporating these state-of-the-art strategies into our approach may lead to further improvements, which could be explored in future work.

\section{Acknowledgment}
This work was supported by the National Natural Science Foundation of China under Grant 62476099 and 62076101, Guangdong Basic and Applied Basic Research Foundation under Grant 2024B1515020082 and 2023A1515010007, the Guangdong Provincial Key Laboratory of Human Digital Twin under Grant 2022B1212010004, the TCL Young Scholars Program, and the 2024 Tencent AI Lab Rhino-Bird Focused Research Program. Dr Tao’s research is partially supported by NTU RSR and Start Up Grants.
\vspace{-8.2mm}
{
    \small
    \bibliographystyle{ieeenat_fullname}
    \bibliography{main}
}
\clearpage
\section*{\centering \LARGE \bfseries Supplementary Material}
\addcontentsline{toc}{section}{Supplementary Material}
In this Supplementary Material, we provide additional experiments, visualization, and illustration to better evaluate our method.

\section{Additional Experiments}
\noindent{\bf Projector for Feature Alignment.}
We conduct experiments to investigate how the number of transformer blocks in the projector affects performance. As shown in Tab.~\ref{tab:blocks}, when the number of blocks is set to 3, our method achieves the best performance, suggesting that increasing the number of blocks in the projector improves the effect of feature alignment. However, excessive complexity in the projector can negatively affect performance.

\begin{table}[htbp]
\caption{Ablation study on the number of blocks in the projector.}
\label{tab:blocks}
\centering
\begin{tabular}{cc}
\hline
\multicolumn{1}{l}{Number of Blocks} & ScanObjectNN \\ \hline
1                                    & 88.34             \\
2                                    & 88.75             \\
3                                    & 90.08        \\
4                                    & 89.03             \\ \hline
\end{tabular}
\end{table}

\noindent{\bf Only Fine-tuning Prediction Head.}
To make a fair comparison with the previous methods, we fine-tune both the backbone and the prediction head in our main experiments. Additionally, freezing the entire 3D backbone and updating only the prediction head can help improve the fine-tuning efficiency, so we conduct experiments and report them in Tab.~\ref{tab:finetune}.

As shown in Tab.~\ref{tab:finetune}, under this configuration, Point-BERT and our method achieve 81.64\% and 85.46\% accuracy, respectively, each requiring one hour of fine-tuning. This approach saves an hour compared to fine-tuning the entire backbone, although it also leads to some performance degradation.

\begin{table}[htbp]
\caption{Ablation study on fine-tuning strategies. Time (h) refers to fine-tuning time on the
PB-T50-RS setting using a single NVIDIA RTX 4090 GPU. 
}
\label{tab:finetune}
\resizebox{\linewidth}{!}{
\begin{tabular}{lcccc}
\hline
\multirow{2}{*}{Method} & \multicolumn{2}{c}{Fine-tuning Whole Backbone} & \multicolumn{2}{c}{Fine-tuning Prediction Head} \\ \cline{2-5} 
                        & Time               & Accuracy              & Time             & Accuracy              \\ \hline
Point-BERT~\cite{yu2022point}              & 2.0                    & 83.07                 & 1.0                    & 81.64                  \\
Ours                    & 2.0                    & 90.08                 & 1.0                    & 85.46                 \\ \hline
\end{tabular}
}
\end{table}

\noindent{\bf Deploying a Pre-trained Encoder in the First Stage.}
To verify whether deploying a pre-trained model in the first stage effectively improves performance, we conduct experiments by loading the Point-BERT weights.

From the results in Tab.~\ref{tab:pretrain}, loading the Point-BERT weights in the first stage yields only slight improvement, as this stage primarily serves to construct the point-to-image framework rather than optimize 3D representations.

\begin{table}[htbp]
\centering
\caption{Ablation study on deploying a pre-trained encoder in the first stage.}
\label{tab:pretrain}
\resizebox{\linewidth}{!}{
\begin{tabular}{lccc}
\hline
\multirow{2}{*}{Method} & \multicolumn{3}{c}{ScanObjectNN} \\ \cline{2-4} 
                        & OBJ-BG  & OBJ-ONLY  & PB-T50-RS  \\ \hline
Point-BERT~\cite{yu2022point}              & 87.43   & 88.12     & 83.07            \\
Ours                    & 95.18   & 93.63     & 90.08           \\
Ours+Point-BERT         & 95.35   & 93.63     & 90.15           \\ \hline
\end{tabular}
}
\end{table}

\noindent{\bf Evaluation on Outdoor Datasets.}
We conduct additional experiments on SemanticKITTI~\cite{behley2019semantickitti} with a SparseConvNet~\cite{graham20183d} backbone to evaluate our method on outdoor LiDAR datasets.

As shown in Tab.~\ref{tab:outdoor}, our method improves the baseline performance from 68.6\% to 69.5\% mIoU, demonstrating its effectiveness in real-world outdoor scenarios. 
\begin{table}[htbp]
\caption{Semantic segmentation results on the SemanticKITTI dataset measured by mIOU (\%).}
\centering
\begin{tabular}{lc}
\hline
Method        & SemanticKITTI \\ \hline
SparseConvNet~\cite{graham20183d} &  68.6             \\
Ours          &  \textbf{69.5}               \\ \hline
\end{tabular}
\label{tab:outdoor}
\end{table}

\noindent{\bf Integration with 3D Intra-modal Self-supervised Loss.}
\begin{table}[t]
\centering
\caption{ Classification accuracy (\%) on the three subsets of the ScanObjectNN dataset and ModelNet40 dataset. }
\resizebox{\linewidth}{!}{
\begin{tabular}{lcccc}
\hline
\multirow{2}{*}{Method} & \multicolumn{3}{c}{ScanObjectNN} & \multirow{2}{*}{ModelNet40} \\ \cline{2-4} 
                       & OBJ-BG  & OBJ-ONLY  & PB-T50-RS  \\ \hline
Point-BERT~\cite{yu2022point}                       & 87.43   & 88.12     & 83.07  & 92.7    \\
Point-FEMAE~\cite{zha2023towards}                       & 95.18   & 93.29     & 90.22   & \textbf{94.0}   \\
ReCon~\cite{qi2023recon}                          & 95.18   & 93.63     & 90.63 & \textbf{94.0}     \\
Ours                             & 95.18   & 93.63     & 90.08 & 93.7     \\
Ours+Point-BERT               & \textbf{95.53}        & \textbf{93.98}          & \textbf{90.67}   & 93.9        \\ \hline
\end{tabular}
}
\label{tab:add_loss}
\end{table}
Our method adopts only a cross-modal alignment loss, while methods like I2P-MAE~\cite{zhang2023learning} and ReCon~\cite{qi2023recon} incorporate both intra-modal and inter-modal self-supervision. This difference in supervision may explain the performance variation observed on some datasets.

To test this hypothesis, we integrate the Point-BERT 3D intra-modal loss into our method in the second stage. As shown in Tab.~\ref{tab:add_loss}, this single-modal self-supervision helps PointSD further improve its performance.

\section{Visualization and Illustration}
\begin{figure}[htbp]
  \centering
  \includegraphics[width=\linewidth]{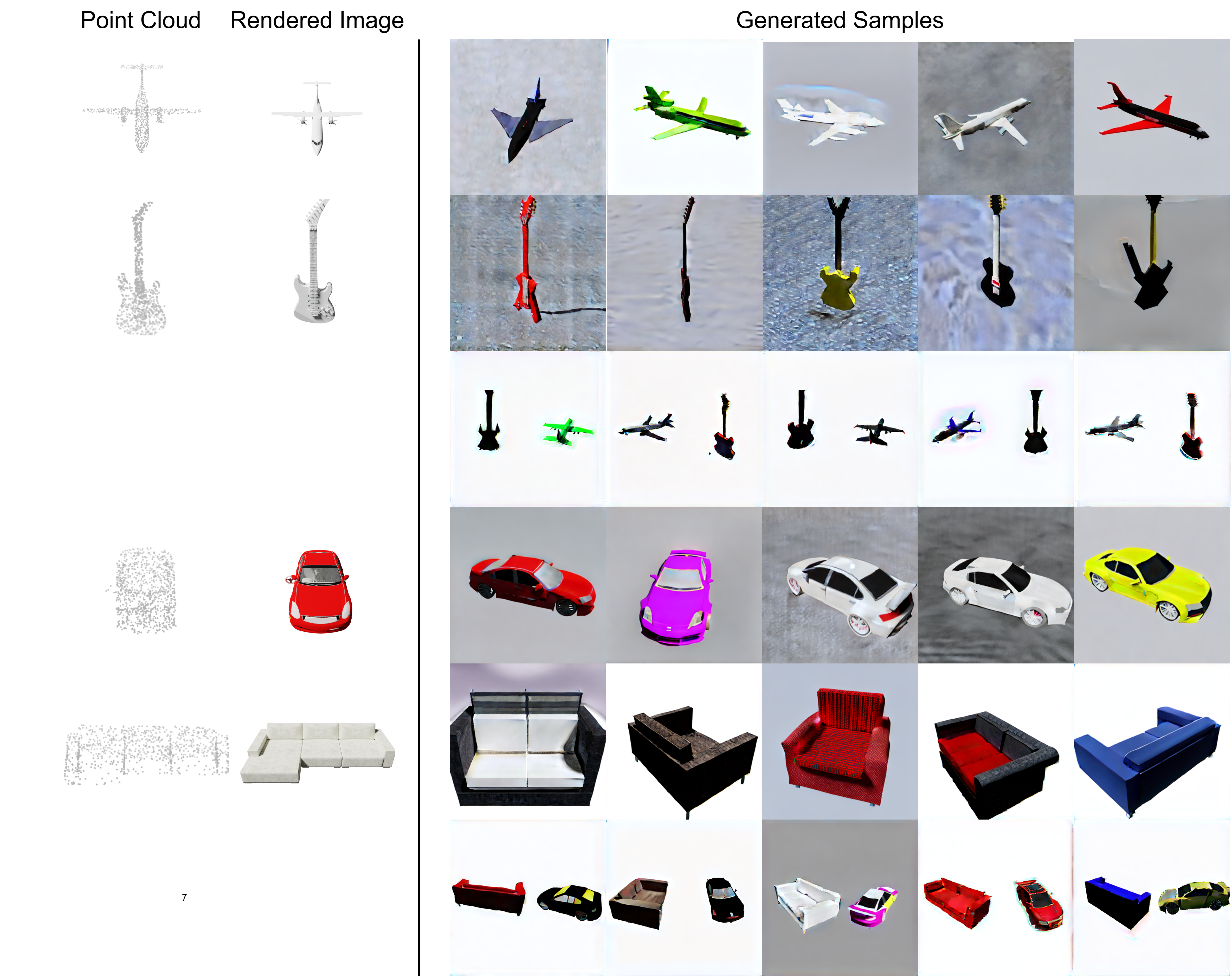} 
      \caption{Visualization of point-to-image generation results. }
      \label{fig:visualization}
      \end{figure}

\noindent{\bf Visualization of Point-to-image Generation.}
As shown in Fig.~\ref{fig:visualization}, we visualize the point cloud and the corresponding rendered image in the left two columns, respectively, and the results generated with different seeds are shown in the five right columns. From top to bottom, the point clouds in the third and sixth rows are generated by mixing the point clouds from rows 1 and 2, and rows 4 and 5, respectively.
The results demonstrate that our point-to-image framework can generate the corresponding images with point clouds as the condition, enabling us to extract SD features containing semantics.
      
\noindent{\bf t-SNE Visualization of Features from 3D Backbone.}
\begin{figure}[htbp]
  \centering
  \includegraphics[width=0.7\linewidth]{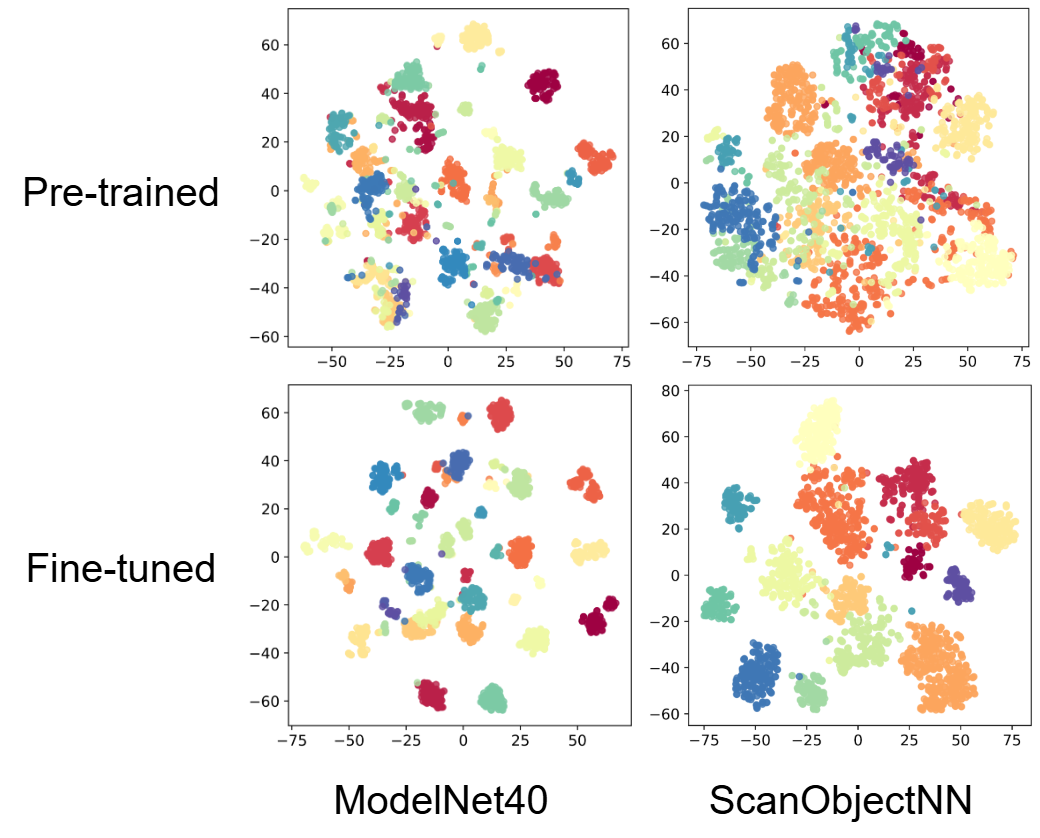} 
      \caption{t-SNE visualization on ModelNet40 and ScanObjectNN PB-T50-RS datasets.}
      \label{fig:tsne}
\end{figure}
In Fig.~\ref{fig:tsne}, we 
achieve t-SNE~\cite{van2008visualizing} visualization on the feature distribution extracted by our pre-trained and fine-tuned models on ModelNet40~\cite{wu20153d} and ScanObjectNN PB-T50-RS~\cite{uy2019revisiting} datasets. The results show that 1) Our pre-trained models can extract discriminative features on the ModelNet40 dataset without fine-tuning. 2) Our fine-tuned models can yield more discriminative features on both datasets. 3) ScanObjectNN PB-T50-RS is a real-world dataset containing background noise, while our model is pre-trained on synthetic data, making it harder for the model to separate different classes of samples in feature space without fine-tuning.

\noindent{\bf Illustration of the Augmentation Strategy.}
\begin{figure}[htbp]
\centering
\includegraphics[width=\linewidth]{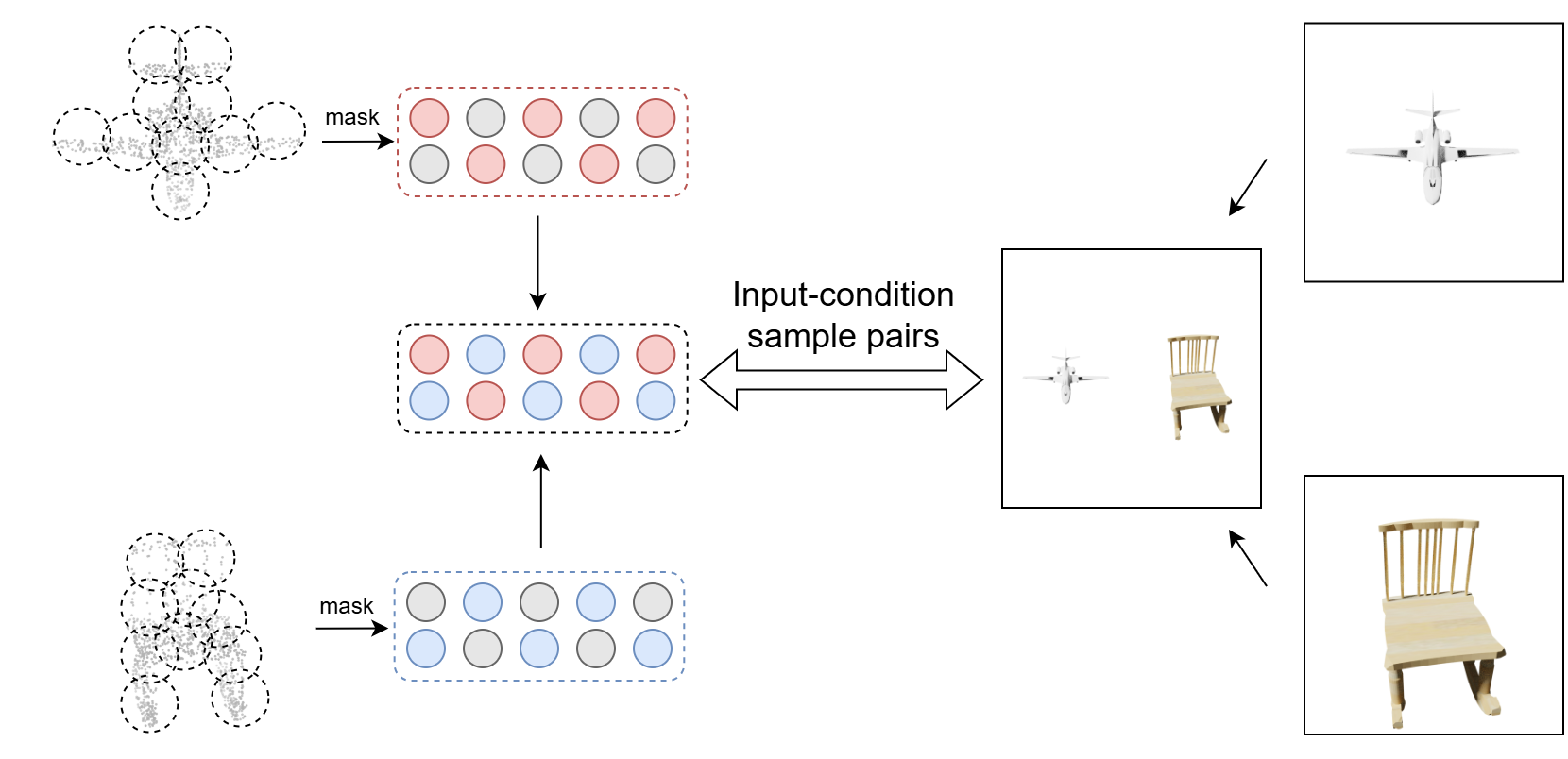}
\caption{Illustration of augmented training samples construction. 
} \label{fig:mix}
\end{figure}
We show our augmentation strategy in Fig.~\ref{fig:mix}. Given two point cloud samples, we first divide them into a series of patches respectively and then mask out part of them to mix. For the corresponding two image samples, we stitch them along the width directly. This augmentation strategy aids in learning more robust 3D representations.

\end{document}